\documentclass[conference]{IEEEtran}
\IEEEoverridecommandlockouts
\usepackage[utf8]{inputenc}
\usepackage{amsmath,amssymb,amsfonts}
\usepackage{bm}
\usepackage{cite}
\usepackage{url}
\usepackage{graphicx,graphics}
\usepackage{colortbl}       
\usepackage{array}
\usepackage{booktabs}
\usepackage{tabularx}
\usepackage[table, svgnames]{xcolor}
\usepackage[flushleft]{threeparttable}
\let\labelindent\relax
\usepackage[inline]{enumitem}
\newcommand{\midsepremove}{\aboverulesep = 0.2mm \belowrulesep = 0.2mm}

\graphicspath{{figures/}}
\DeclareGraphicsExtensions{.pdf,.jpeg,.png,.tiff,.jpg}

\title{Haptic-guided assisted telemanipulation approach for grasping desired objects from heaps
\thanks{This work was supported by the UK National Centre for Nuclear Robotics (NCNR). Part funded by EPSRC EP/S032428/1 and EP/P01366X/1 grants.}
}
\author{\IEEEauthorblockN{Maxime Adjigble}
\IEEEauthorblockA{\textit{Extreme Robotics Laboratory} \\
\textit{University of Birmingham}\\
Birmingham, UK \\
m.k.j.adjigble@bham.ac.uk}
\and
\IEEEauthorblockN{Rustam Stolkin}
\IEEEauthorblockA{\textit{Extreme Robotics Laboratory} \\
\textit{University of Birmingham}\\
Birmingham, UK \\
r.stolkin@bham.ac.uk}
\and
\IEEEauthorblockN{Naresh Marturi}
\IEEEauthorblockA{\textit{Extreme Robotics Laboratory} \\
\textit{University of Birmingham}\\
Birmingham, UK \\
n.marturi@bham.ac.uk}
}
\begin{document}\sloppy
\bstctlcite{IEEEexample:BSTcontrol}
\maketitle
\begin{abstract}
This paper presents an assisted telemanipulation framework for reaching and grasping desired objects from clutter. Specifically, the developed system allows an operator to select an object from a cluttered heap and effortlessly grasp it, with the system assisting in selecting the best grasp and guiding the operator to reach it. To this end, we propose an object pose estimation scheme, a dynamic grasp re-ranking strategy, and a reach-to-grasp hybrid force/position trajectory guidance controller. We integrate them, along with our previous SpectGRASP grasp planner, into a classical bilateral teleoperation system that allows to control the robot using a haptic device while providing force feedback to the operator. For a user-selected object, our system first identifies the object in the heap and estimates its full six degrees of freedom (DoF) pose. Then, SpectGRASP generates a set of ordered, collision-free grasps for this object. Based on the current location of the robot gripper, the proposed grasp re-ranking strategy dynamically updates the best grasp. In assisted mode, the hybrid controller generates a zero force-torque path along the reach-to-grasp trajectory while automatically controlling the orientation of the robot. We conducted real-world experiments using a haptic device and a 7-DoF cobot with a 2-finger gripper to validate individual components of our telemanipulation system and its overall functionality. Obtained results demonstrate the effectiveness of our system in assisting humans to clear cluttered scenes.
\end{abstract}
\begin{IEEEkeywords}
Shared control, haptic systems, grasping
\end{IEEEkeywords}
%
\section{Introduction}
\label{sec:intro}
Human-in-the-loop robotic telemanipulation integrates human expertise and robot capabilities to achieve higher efficiency in performing complex tasks. It has gained significant attention in recent years as a promising solution to improve the safety of human-robot collaboration. These systems are widely used to perform tasks such as decommissioning hazardous waste in nuclear sites \cite{canbolat2017robots,marturi2016towards}, performing invasive surgeries \cite{haidegger2022robot}, exploring deep oceans and outer space \cite{jakuba2018teleoperation}, and conducting search and rescue missions \cite{7379322}. While modern robotic arms can perform repetitive tasks with higher precision and speed than human workers, more intricate tasks requiring fine manipulation and decision-making skills still require human input. Despite the benefits of improved safety and greater efficiency, performing remote telemanipulation using a multi-degrees of freedom (DoF) robotic arm (with a joystick device) is challenging due to the lack of feedback, complexity in controlling multi-DoF robots, and limited task space with a lack of depth perception. For instance, to clear a remote scene cluttered with various objects heaped together, operators must identify suitable poses to grasp the objects and make informed decisions on driving the robot end-effector (controlling both position and orientation) to those locations while avoiding collisions and robot singularities. 

Since the introduction of virtual fixtures in the 90s \cite{rosenberg1993virtual}, numerous assistance/haptic guidance systems have been proposed in the literature to perform a variety of tasks, with the majority focused on surgical applications. In this paper, we specifically discuss the works that provide operator guidance while grasping objects in a workspace. In \cite{ghalamzan2017human}, a haptic-based shared control method was presented to assist users in driving the robot towards the best grasping pose that maximizes manipulability. In \cite{laghi2022target}, a shared autonomy method was presented to guide operators in reaching and manipulating box-shaped objects with one or two arms (in a bimanual setup), depending on the size of the box. This approach also uses visual cues to understand operator intention and automatically adapt robot trajectories. In \cite{abi2016visual}, a haptic-based shared controller was presented to approach and grasp an object. Here, the gripper orientation is constrained that it is always oriented towards the object. Although all these approaches demonstrated good performance, they are limited to working with a single object at a time, as the grasps are pre-computed. 
Addressing these limitations, authors in \cite{abi2019haptic} presented a shared control approach to work with multi-object scenes. In \cite{adjigble2019assisted}, an automatic grasp selection approach was presented that can pick-up objects from clutter. With both these methods, grasps are computed for entire scene, \textit{i.e.}, for all the objects in the scene. Further, a dynamic re-ranking scheme is used in \cite{adjigble2019assisted} to update the feasible grasps based on the end-effector position. However, end-effector orientations are not considered, which we consider in this paper.

In this paper, \textbf{we propose an intuitive assisted telemanipulation architecture} that allows human operators to perform pick and place various objects from a heap. With this system, the operator will be able to define high-level goals, such as selecting the object to handle and determining the approach direction for the robot, while an autonomous agent handles the object's grasp position and feasible trajectory to the target. For a cluttered scene, our system first generates the scene point cloud by registering camera-acquired point clouds from multiple viewpoints. The operator selects an object to pick through the provided terminal-based interface. In this work, the list of objects constituting the clutter are known a priori and their CAD models are available beforehand. First, the user selected object is identified in the scene point cloud and its pose is estimated. \textbf{We propose a spectral domain-based pose estimation method}, which maps the reference model of the selected object onto the scene point cloud. Using the estimated object pose and the corresponding point cloud, our learning-free grasping algorithm, SpectGRASP \cite{adjigble2021spectgrasp}, generates all the feasible collision-free grasp candidates for the object. In contrast to our previous work \cite{adjigble2019assisted}, where the grasp candidates are estimated for entire scene, the proposed system now estimate feasible grasps only for the selected object. \textbf{We integrate grasp planning with force guidance for telemanipulation}, where natural hand movements (via haptic device) are coupled with the robot movements. During assisted telemanipulation, the top-ranked grasp is used as the target pose by the shared controller, which generates a feasible trajectory for the robot to follow with haptic feedback. It is worth noting that the ranking module within the grasp planner automatically ranks the generated grasp candidates at the time of generation. Further, the top-ranked grasp is dynamically updated based on the current position of the robot end-effector so as to ensure that the generated assisted trajectory is kinematically feasible for the robot to move from its current location. To this extent, \textbf{we propose a new re-ranking schema}, which in contrast to our previous work \cite{adjigble2019assisted}, takes into account both position and orientation of the robot end-effector. Finally, \textbf{we integrated a shared pose controller to automatically handle the end-effector orientations} while reaching to grasp, \textit{i.e.}, during assisted teleoperation, robot position control is performed by human operator while the orientations are automatically handled by the autonomous agent. Note that the human operator can turn on and off the force guidance from our haptic interface. We demonstrate the efficiency of our system by conducting multiple randomly generated real-world clutter clearance experiments using a 7-DoF collaborative robot fixed with a parallel-jaw gripper.
%
%
%
\section{Methodology}
\label{sec:method}
In this section, we present our human-in-the-loop assisted telemanipulation approach to remove objects from heap. Our method's pipeline consists of the following steps: 
\begin{enumerate*}[label=(\roman*)]
  \item acquire the scene point cloud; 
  \item identify and segment the user-selected object from the scene cloud;
  \item plan grasps on the identified object;
  \item teleoperate the robot using a haptic device to reach grasp pose;
  \item re-rank the grasps as the robot moves;
  \item when force guidance is activated by the operator, provide virtual haptic force feedback, and orientation shared control to reach and grasp the selected object.
\end{enumerate*}
In the following sections, we present solutions to the problems associated with these steps.
\subsection{Point cloud representations} \label{subsec:pc_representation}
Earlier, we mentioned that point clouds are used in this work. Additionally, we take into account their point surface normals. When using normals, there are multiple representations available. In this work, we consider two representations: Extended Gaussian Images (EGI) and Binary Extended Gaussian Images (BEGI). EGI provides a histogram representation of the surface normals on the unit sphere $\bm{S}^2$, which offers more information about an object's shape than BEGI, which only contains binary information on the normal orientations. Therefore, we use EGIs for pose estimation and BEGI for grasping. This representation has previously proven to be useful for many tasks \cite{adjigble20233d,adjigble2021spectgrasp}.

Let $\mathcal{P}_c$ be a point cloud consisting of $N \in \mathbb{N}^+$ points with coordinates $p_i=(x_i,y_i,z_i)$ and surface normals  $n_i=(n^i_x,n^i_y,n^i_z)$, with $i \in [1, N]$. While normals are represented as EGI, points are represented as a voxel grid. In the voxel grid, each voxel contains binary values that indicate the presence of points in the cell. Alternatively, a voxel grid with real values could be used, but this would require encoding the voxel's 3D points into a single real value that may provide information about the local surface of the object. Other potential candidates for this representation include curvatures (minimum, maximum, or a combination of both) or the Local Contact Moment score (LoCoMo) \cite{adjigble2018model}. However, these options are not investigated in this work and are left for future studies. %
Given a resolution $\mathcal{R} \in \mathbb{R}^+$ and the voxel indices $p_{ijk}= (i,j,k)$, the voxel value function $f_t(p) = f_t(x, y, z) = v_{ijk} \in [0,1]$ is expressed as%
%
%
\begin{equation}
\label{eq:voxel_index}
i = [x/\mathcal{R}]\qquad
    j = [y/\mathcal{R}]\qquad
    k = [z/\mathcal{R}]
\end{equation}
\begin{equation}
\label{eq:voxel_value}
v_{ijk}= 
\begin{cases}
    1,& \text{if a point with indices i,j,k exists} \\
    0,              & \text{otherwise}
\end{cases}
\end{equation}
The operation $[./.]$ is the integer division, 
and $t$ in subscript indicates that the function is used for estimating translation.

\begin{figure}
    \centering
    \includegraphics[width=\columnwidth]{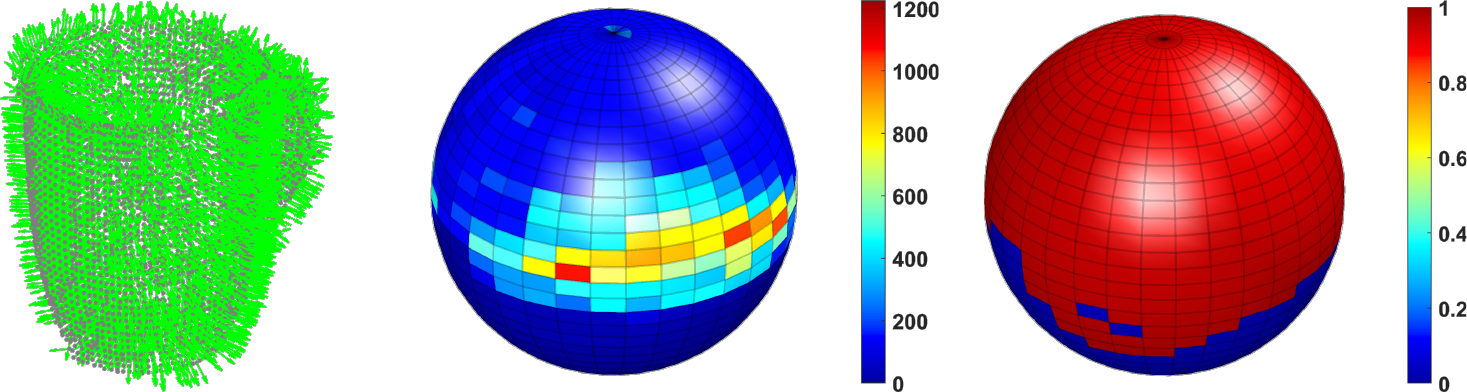}
    \caption{(middle) EGI and (right) BEGI of a point normal cloud.}
    \label{fig:egi_begi}
\end{figure}

To compute the EGI, the coordinates of surface normals $n \in \mathbb{R}^3$ must first be converted from Cartesian to spherical coordinates, i.e., $n = (1, \theta, \phi)$. Since the radial distance is equal to 1 (surface normals are unit vectors), $n$ can be described by the set $(\theta, \phi)$. Given a bandwidth $B = 2^b, b \in \mathbb{N}^+$, the unit sphere can be discretized along the longitude and the latitude using $\theta_j = \frac{\pi(2j + 1)}{4B}$ and $\phi_k = \frac{\pi k}{B}$, where $(j, k) \in \mathbb{N}$ subject to the constraint $0 \leq j, k < 2B$ on the indices. The EGI of the surface normals can be expressed at each $(\theta_j, \phi_k)$ by the function $f_r : \bm{S}^2 \rightarrow \mathbb{N}$:
\begin{equation}
f_r(\theta_j, \phi_k) = c_{j,k}
\label{eq:egi}
\end{equation}
where, $c_{j,k}$ is the count of normals in the cell $(\theta_j, \phi_k)$. %
Similar to EGI computation, the BEGI of a point cloud is computed by limiting the maximum value of $c_{j,k}$ in \eqref{eq:egi} to $1$ and additionally storing the Cartesian coordinate $p_i$ of the points in a point set map $\mathcal{P}_{jk}$ function. 
Thus, the BEGI of a point cloud $g_{br}$ is defined as:
\begin{equation}\label{eq:begi_representation}
\begin{gathered}
    g_{br}(\theta_j, \phi_k)  = 
\begin{cases}
    1,& \text{if } c_{j,k} > 0\\
    0,              & \text{otherwise}
\end{cases} \\
   \mathcal{P}_{jk} = \left\{ p_i \in  \mathbb{R}^3 \mid n_i = (\theta_j, \phi_k) \right\}
\end{gathered}
\end{equation}
Fig. \ref{fig:egi_begi} shows EGI and BEGI for a point normal cloud.
%
%

\subsection{Fourier transforms on $\mathbb{R}^3$ and $\bm{SO}(3)$}
The object poses in this work are computed using Fourier analysis. Therefore, in this section, we will provide a brief overview of the Fourier transformations necessary for the subsequent sections. The Fourier transform is a widely studied and crucial tool in signal processing and pattern recognition \cite{groemer1996geometric}. Its primary advantage is that operations such as convolution and correlations, which assess the similarity between two signals, can be efficiently carried out in the frequency domain. This property has enabled the development of highly effective algorithms. Leveraging this fact, we propose a new 3D object pose estimator that aligns a reference model of an object with a scene cloud.%

\subsubsection{Fourier transform on $\mathbb{R}^3$}
With $f_t: \mathbb{R}^3 \rightarrow \mathbb{N}$ being the voxel value function of a point cloud of an object or scene, the Fourier transform of $f_t$ is computed by:
\begin{equation}
F_t(u, v, w) = \sum_{x=0}^{M-1}\sum_{y=0}^{N-1}\sum_{z=0}^{L-1}f_t(x,y,z)e^{-i2\pi(\frac{u}{M}x + \frac{v}{N}y + \frac{w}{L}z)}
\label{eq:fourier_coefs}
\end{equation}
where,  $F_t(u, v, w)$ is the Fourier coefficient evaluated at frequency $(u, v, w)$ and $M, N, L \in \mathbb{N}^+$ are the number of samples in the $X$, $Y$, and $Z$ directions, respectively. The algorithms to efficiently compute these Fourier coefficients are available in the literature \cite{brigham1988fast}.

\subsubsection{Fourier transform on $\bm{SO}(3)$}
Given a function $f: \bm{S}^2 \rightarrow \mathbb{N}$, which can be either an EGI or a BEGI, the following Fourier decomposition holds:
\begin{equation}
f(\theta, \phi) = \sum_{l=0}^{l_{max}}\sum_{m=-l}^{l}\hat{f_l^m}Y_l^m(\theta, \phi)
\label{eq:spherical_harmonics}
\end{equation}
where,   $Y_l^m$ is the spherical harmonics of order $m$ and degree $l$, with $l, m \in \mathbb{N}^+$ .  $l_{max} \in \mathbb{N}^+$ is the maximum degree of expansion of the series, $\hat{f_l^m}$ is the harmonic coefficient. $\hat{f_l^m}$ is computed by integrating the inner product of $f_r$  and $\overline{Y_l^m(w)}$, the complex conjugate of $Y_l^m$, over the unit-sphere $\bm{S}^2$ . It is written as follows:
\begin{equation}
\label{eq:harmonic_coeffs}
\hat{f_l^m} = \int_{w \in \bm{S}^2} f(w)\overline{Y_l^m(w)} \,dw 
\end{equation}
A method to compute harmonic coefficients for any complex valued square-integrable function on $\bm{S}^2$ is presented in \cite{kostelec2008ffts}.

\subsection{Object 3D pose estimation} \label{sec:pose_estimation}
Given a scene and a known reference object model, the problem of pose estimation can be formulated as finding the optimal transformation $H_{opt} \in \mathbb{R}^3 \times \bm{SO}(3)$ between the reference object and its instance in the scene. This problem has been extensively studied in the literature \cite{huang2021comprehensive}. The scene may also be partially visible and consist of various objects with different shapes, which further complicates the problem. In this paper, we formulate the pose estimation problem as finding the location of the peak correlation between the point cloud of the scene and reference model on $\mathbb{R}^3 \times \bm{SO}(3)$. However, directly computing the correlation function over $\mathbb{R}^3 \times \bm{SO}(3)$ is a challenging task. To overcome this challenge, we propose a two-step approach. First, we estimate potential 3D rotations by computing the correlation $C_r$ of the EGIs of the scene and reference model over $\bm{SO}(3)$. Next, for each rotation candidate, we compute the correlation $C_t$ of the voxel grids of the scene and reference model over $\mathbb{R}^3$ to estimate the translation. The resulting candidates are then ranked based on the value of $C_t$, with the highest value indicating the most probable object pose 
\subsubsection{Sampling rotation candidates}
Let $f_r$ and $g_r$ be the EGIs of the scene and reference object models, respectively and $\bm{R} \in \bm{\mathrm{SO}(3)}$ be a rotation parametrised by the $zyz$ Euler angles $(\alpha, \beta, \gamma)$. The correlation $C_r(\bm{R})$ of  the two clouds can be computed by integrating the inner product of $f_r$ and the complex conjugate of $g_r$ over $\bm{S}^2$:
\begin{equation}
    \mathcal{C}_r(\bm{R}) = \int_{w \in S^2} f_r(w)\overline{g_r(w)} \,dw
    \label{eq:correlation_so3}
\end{equation}
As presented in \cite{kostelec2008ffts}, the previous equation could further be simplified by discretising the $\bm{S}^2$ space, using the Fourier transforms of the functions and applying the orthogonality principle of the spherical harmonics. It is then re-written as: 
\begin{equation}
    \mathcal{C}_r(\bm{R}) = \sum_{l=0}^{l_{max}}\sum_{m=-l}^{l}\sum_{m'=-l}^{l} \hat{f_l^m}\overline{\hat{g_l^{m'}}} \overline{D_{mm'}^l(\bm{R})} 
    \label{eq:correlation_so3_simplified}
\end{equation}
where, $D_{mm'}^l$ is the Wigner D-matrix. \eqref{eq:correlation_so3_simplified} can be used to efficiently evaluate the correlation between $f_r$ and $g_r$. Next, a correlation map is computed by evaluating $\mathcal{C}_r(\bm{R})$ at a set of discrete Euler angles values. Sampling rotations from the correlation map for which $\mathcal{C}_r(\bm{R})$ is greater than a predefined threshold $tc_r$ provides the set of potential rotation candidates:
\begin{equation}\label{eq:rotation_candidates}
   \mathcal{R}_{tc_r} = \left\{ R \in  \bm{SO}(3) \mid \mathcal{C}_r(\bm{R}) > tc_r \right\}
\end{equation}
\subsubsection{Estimate the object pose}
Let $f_t$ and $g_t$ be the voxel grids of the scene and reference object, respectively. 
For $\bm{R} \in \mathcal{R}_{tc_r}$, the Fourier shift property can be used to find the optimal Cartesian translation $T_{opt}(\bm{R})$ between the rotated object model and the scene. Let  $g_t^R = g_t(\bm{R})$ be the voxel grid of the object model rotated by $\bm{R}$. The optimal translation is found by computing the inverse Fourier transform of the normalised cross-power spectrum $\mathcal{C}_t$ of $f_t$ and $g_t^R$:
\begin{equation}
\begin{aligned}
\mathcal{C}_t(u,v,w) &= \frac{F_t(u,v,w)\overline{G_t^R(u,v,w)}}{|F_t(u,v,w)\overline{G_t^R(u,v,w)}|} \\
\delta(T) &= \mathcal{F}^{-1}(\mathcal{C}_t(u,v,w)) 
\label{eq:cross_power_spectrum}
\end{aligned}
\end{equation}
where,  $F_t$ and $G_t^R$ are the Fourier coefficients of $f_t$ and $g_t^R$, respectively. $\mathcal{F}^{-1}$ is the inverse Fourier transform. $\delta(T)$ is the Dirac Delta function whose peak location corresponds to the optimal translation $T_{opt}(\bm{R})$. The Dirac peak value $\delta_{max}$ indicates the degree of correlation between the transformed object model and the scene. Higher correlations are desirable because they indicate a greater overlap between the transformed object model and the scene. The optimal transformation $H_{opt}$ can then be found by: 
\begin{equation}\label{eq:optimal_transform}
   H_{opt} = \max_{\delta_{max}}\left\{ (T_{opt}(\bm{R}), \bm{R}), \bm{R}  \in  \mathcal{R}_{tc_r} \right\}
\end{equation}
Each transformation in the set is ranked based on the decreasing values of $\delta_{max}$, where top $K$ could be used as transformation candidates.
%
%
\subsection{Grasp generation}\label{subsec:grasp_generation}
As mentioned earlier, we have utilised our previously developed SpectGRASP method \cite{adjigble2021spectgrasp} to generate grasp candidates. 
This method, which also utilises Fourier transformations on $\bm{SO}(3)$, is capable of generating grasps for single and multi-object scenes. We briefly present the method below. %

A grasp is defined by the set of points and normal vectors $(p_i, n_i)$ corresponding to the location of the contacts between the robot hand fingers and an object, and the wrist pose of the robot hand $H_g \in \mathbb{R}^3 \times \bm{SO}(3)$. %
The problem of grasping is then to find the set of grasps $\bm{\mathcal{G}}$ that produce a high correlation $\mathcal{C}_t$ between the robot fingers and the scene:
\begin{equation}
    \bm{\mathcal{G}} = \{(p_1\cdots p_{N_f}, n_1\cdots n_{N_f}, H_g)  \mid \mathcal{C}_t > tc_g\}
    \label{eq:grasp_conf}
\end{equation}
where, $N_f$ is the number of fingers of the robot hand and $tc_g$ is a threshold. It is important to note that not all fingers need to be in contact with the objects. However, defining a grasp in this manner constrains the position of all the fingers, including those not in contact. The robot hand finger geometries can be discretised and converted to a point cloud with surface normals. 
In this case, the previously presented pose estimation method could be used to sample robot hand poses on the surface of objects. However, since the gripper can have multiple DoFs, the pose estimation method (using EGIs) is impractical as it would require performing the estimation for each joint angle of the robot hand. As a result, we opt to use BEGI to address the grasping problem. 

Given a robot hand configuration specified by its joint angles $q$, robot hand orientations are sampled using \eqref{eq:rotation_candidates}. This allows us to identify scene points for which the surface normals are oriented in the same way as those of the robot hand for the given hand configuration. These points are obtained by rotating the robot hand for each rotation in $\mathcal{R}_{tc_r}$ and extracting the points from the scene's BEGI where each finger surface normal falls. The force closure principle \cite{nguyen1988constructing} is then used on the set of extracted points to filter out unstable grasps. The remaining grasps are ranked using LoCoMo metric \cite{adjigble2018model}. Wrist poses $H_g$ are sampled by using the kinematics of the robot hand.  In this work, a parallel-jaw gripper is used, which means that the orientation of the surface normals of the fingers is independent of the value of the joint configuration. This means that all the robot hand configurations are considered using the presented approach.

\subsection{Bilateral haptic teleoperation} \label{sec:bilateral_teleop}
\begin{figure}
    \centering
    \includegraphics[width=\columnwidth]{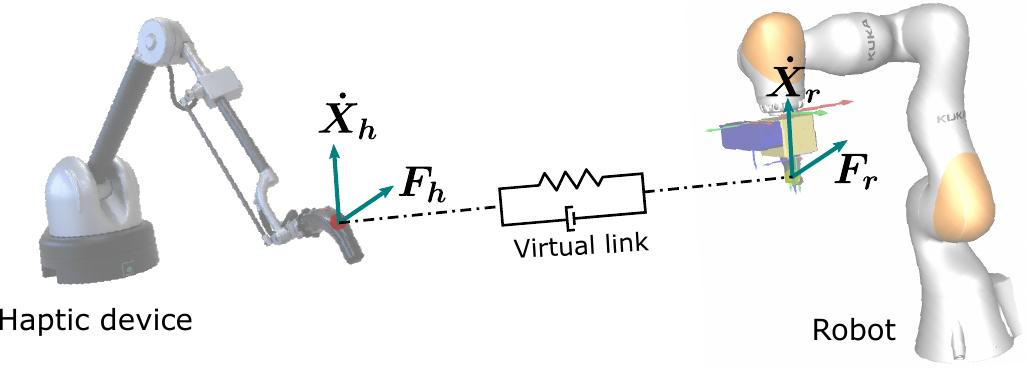}
    \caption{Illustration of bilateral coupling for haptic teleoperation.}
    \label{fig:bilat_coupl}
\end{figure}

Bilateral teleoperation allows an operator to control a robot using a haptic device and receive force feedback as the robot interacts with the environment. A significant amount of research has been conducted on this topic in the literature, and different methods have been proposed to accomplish the intended behavior. The most prevalent approach is to simulate a spring and damper system virtually, connecting the end-effectors of the robot and the haptic device as shown in Fig. \ref{fig:bilat_coupl}. This facilitates the movement of the robot as the haptic device moves and vice versa.
Assuming $\bm{X}_r$ and $\bm{X}_h$ be the Cartesian positions of the robot and haptic device, the joint forces $\bm{\tau}_r$ and $\bm{\tau}_h$ required for bilateral coupling can be calculated as follows:
\begin{equation} 
\label{eq:teleop}
\begin{aligned} 
\bm{F}_{r} &= K_{p}(\bm{X}_{h} - \bm{X}_{r}) + K_{d}(\dot{\bm{X}}_{h} - \dot{\bm{X}}_{r})  \\
\bm{\tau}_{r}& = \bm{J}_r^{T}\bm{F}_{r}~, \qquad \bm{\tau}_{h} = -\bm{J}_h^{T}\bm{F}_{r}
\end{aligned} 
\end{equation}
where, $K_{p}$, $K_{d}$, $\bm{J}_r^{T}$ and $\bm{J}_h^{T}$ are the virtual stiffness, damping gain, Jacobian transpose of the robot and haptic device, respectively. $\bm{F}_r$ is the force applied to the robot.

\subsection{Grasp Re-ranking}\label{subsec:reranking}
As the user moves the robot, the grasps are re-ranked on the fly. 
In this work, we propose a re-ranking strategy based on the dual quaternion error between the current pose of the robot hand and the grasp candidates. 
The error is then used to adjust the initial ranking of the grasps. The closest, top-ranked grasp is selected as the best grasp. Since the metric incorporates the dual quaternion error, orientation errors are also taken into account. 

Let $H_{hand}$ be the homogeneous matrix representing the current pose of the robot hand and $\bm{q}_{hand}$ is its dual quaternion representation. Let $g_i = (\bm{p}, \bm{n}, \bm{q}_{g_i})$ be a grasp from $\bm{\mathcal{G}}$, with $r_i$ being its LoCoMo ranking score. Here, $\bm{p}$, $\bm{n}$, and $\bm{q}_{g_i}$ represent the contact point set, corresponding surface normal set, and the dual quaternion representing the pose of the wrist, respectively.  Let us introduce the following symbols:
%
\begin{equation}
\label{eq:dq_error}
\begin{aligned}
\bm{q}_e = \bm{q}_{g_i}^*\bm{q}_{hand} \qquad 
\bm{t} = trans(\bm{q}_e) \\
\bm{\hat{l}} = axis(\bm{q}_e) \qquad
\theta = angle(\bm{q}_e) \\
\end{aligned}
\end{equation} 
The operation $(.)^*$ represents dual-quaternion conjugate and $trans(.)$ extracts the translation from the dual quaternion. For a dual quaternion defined as $\bm{q} = p + \epsilon q$, we can obtain the translation as $\bm{t} = 2qp^*$. The functions $angle(.)$ and $axis(.)$ extracts the angle-axis representation of the rotational part of the dual quaternion. 
More details on this derivation can be found in \cite{jia2013dual}. The updated ranking score $r_i^{'}$ can then be computed as:
\begin{equation}
\label{eq:re_ranking}
\begin{aligned}
d_r &= (\lambda_{t}\bm{t}.\bm{\hat{l}})^2 + (\lambda_{r}\theta)^2\\
r_i^{'} &= \frac{max_{\bm{\mathcal{G}}}(d_r) - d_r}{max_{\bm{\mathcal{G}}}(d_r) - min_{\bm{\mathcal{G}}}(d_r)}r_i
\end{aligned}
\end{equation} 
where, $\lambda_{t}$ and $\lambda_{r}$ are normalising terms. 
$\lambda_{r}$ is set to $1/\pi$, which is the inverse of the maximum value possible for $\theta$, and  $\lambda_{t}$ is set to the inverse of the maximum computed translation distance in $\bm{\mathcal{G}}$. 
$min_{\mathcal{G}}(d_r)$ and $max_{\mathcal{G}}(d_r)$ are the minimum and maximum distances $d_r$ from all the grasps in $\bm{\mathcal{G}}$. $d_r$ serves as a distance metric between the current robot hand and grasp poses. It is utilized to increase the ranking score of grasps that are closer to the current end-effector pose of the robot, \textit{i.e.}, in terms of both position and orientation.

\subsection{Haptic virtual force guidance}\label{subsec:force_guidance}
At each stage of operation, the current position of the robot hand and the closest-best grasp are known. When the force guidance is requested by the operator, a collision-free path is computed between those positions. While the robot could autonomously execute the trajectory and grasp the object, safety-critical applications require a human in the loop at all times. Our approach utilizes a hybrid controller, where a virtual force is applied to the haptic device, allowing the operator to move the robot along the computed trajectory while the robot's orientation is automatically interpolated based on its current position on the path. The force, $\bm{F}_r^*$, required to bring the robot back to the zero-force-torque trajectory is:
%
%
\begin{equation} 
\label{eq:traj_force}
\bm{F}_r^* = K_p^*(\bm{X}_r^* -\bm{X}_r) + K_d^*(\dot{\bm{X}}_r^* - \dot{\bm{X}}_r)
\end{equation}
where, $K_p^*$ and $K_d^*$ are stiffness and damping gains, respectively. $\bm{X}_r^*$ is the closest trajectory pose to $\bm{X}_r$, which is computed by discretising the trajectory and finding the closest pose to $\bm{X}_r$. 
$\dot{\bm{X}}_r$ is the derivative of $\bm{X}_r$. $\bm{F}_r^*$ is overlaid on the bilateral haptic force $\bm{F}_r$ given in \eqref{eq:teleop} in a specific way to enable the desired behaviour. Given a Cartesian force $\bm{F}$, $\bm{F}^F$ and $\bm{F}^{T}$ denote its $x-y-z$ force and torque vectors, respectively. The joint torques applied to the robot and haptic device in case of assisted mode are computed as
\begin{equation} 
\label{eq:teleop_2}
\bm{\tau}_{r} = \bm{J}_r^{T}(\bm{F}_{r}^F + \bm{F}_r^{*T} )\quad
\bm{\tau}_{h} = -\bm{J}_h^{T}(\bm{F}_{r}^F + \bm{F}_r^{*F} )
\end{equation}
The orientation of the robot is controlled by $\bm{F}_r^{*T}$, while its position is exclusively controlled by $\bm{F}_{r}^F$, i.e., contribution of the human presented in the loop. Only the force components of $\bm{F}_{r}$ and $\bm{F}_r^*$ are used to compute the virtual force feedback, providing complete freedom in the orientation of the haptic device. This is necessary to release constraints on the haptic device's orientation, which otherwise can result in uncomfortable jogging positions for the operator.
%
\section{Experimental Validations}
\label{sec:results}
We conducted several experiments to evaluate the effectiveness of our proposed telemanipulation method in clearing cluttered scenes. In addition, we performed qualitative analysis to assess the feasibility of each component of our method's architecture. In this section, we first present the experimental setup and then discuss the experimental results.
\subsection{Experimental setup}
\begin{figure}
    \centering
    \includegraphics[width=0.95\columnwidth]{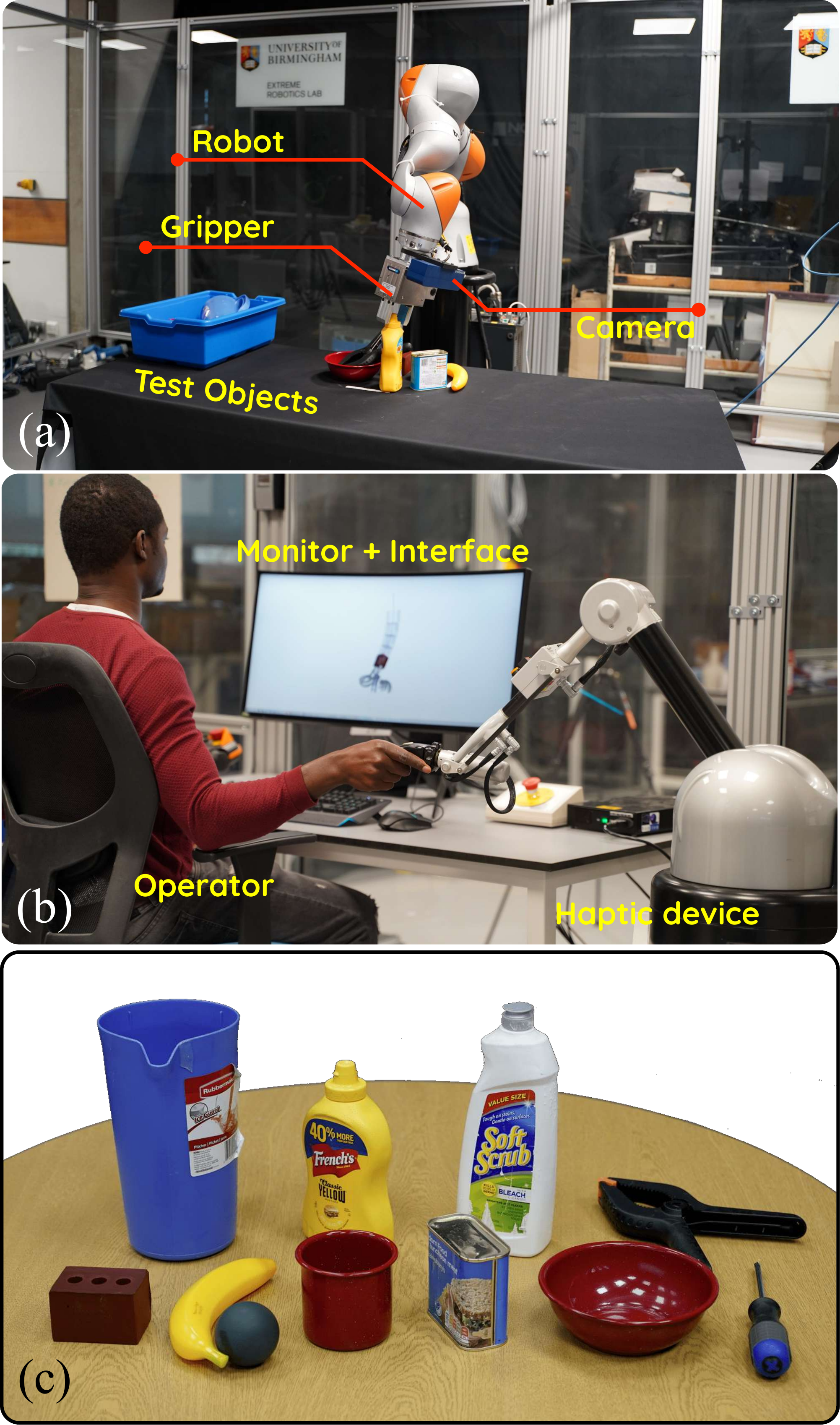}
    \caption{Experimental setup: (a) robot side, (b) operator side. (c) Used test objects from the YCB object set \cite{ycb}.}
    \label{fig:setup_objects}
\end{figure}
Fig. \ref{fig:setup_objects} shows the developed experimental setup. It consists of two sides: robot and operator. On the robot side, a 7-DoF KUKA iiwa robot fixed with a Schunk PG70 two-finger gripper (Fig. \ref{fig:setup_objects}(a)) is used. A 3D camera, Ensenso N35, is mounted on top of the gripper to perceive the environment. To maximise the grasp success, we use full view of the scene instead of a single partial view. The camera is moved to 4 different locations around the scene and the point clouds acquired at these locations are stitched together to form the complete scene. This stitching process is straightforward as the precise location of the camera with respect to the robot base is known. The points belonging to the table are automatically cropped out at the time of acquisition. On the operator side (Fig. \ref{fig:setup_objects}(b)), a Haption Virtuose 6D 6-axis (three translations and three rotations) haptic device is used, along with a monitor featuring a terminal-based interface for the operator to select the objects to handle. Note that the proposed framework is not limited to this setup, and is compatible with any robotic telemanipulation system with a 3D camera and a haptic device.

For validation purposes, we used 11 objects (Fig. \ref{fig:setup_objects}b) from the YCB objectset \cite{ycb}. 3D models of these objects are downloaded from the YCB website and converted to point clouds with surface normals. These point clouds serve as reference models for the pose estimation module.  A list of these objects is provided to the operator, who can select the object to grasp by inputting its ID in the terminal-based interface. The operator can monitor the process in the provided visualisation window.
\subsection{Fundamental component analysis}
\subsubsection{Pose estimation analysis}
\begin{figure}
    \centering
    \includegraphics[width=0.95\columnwidth]{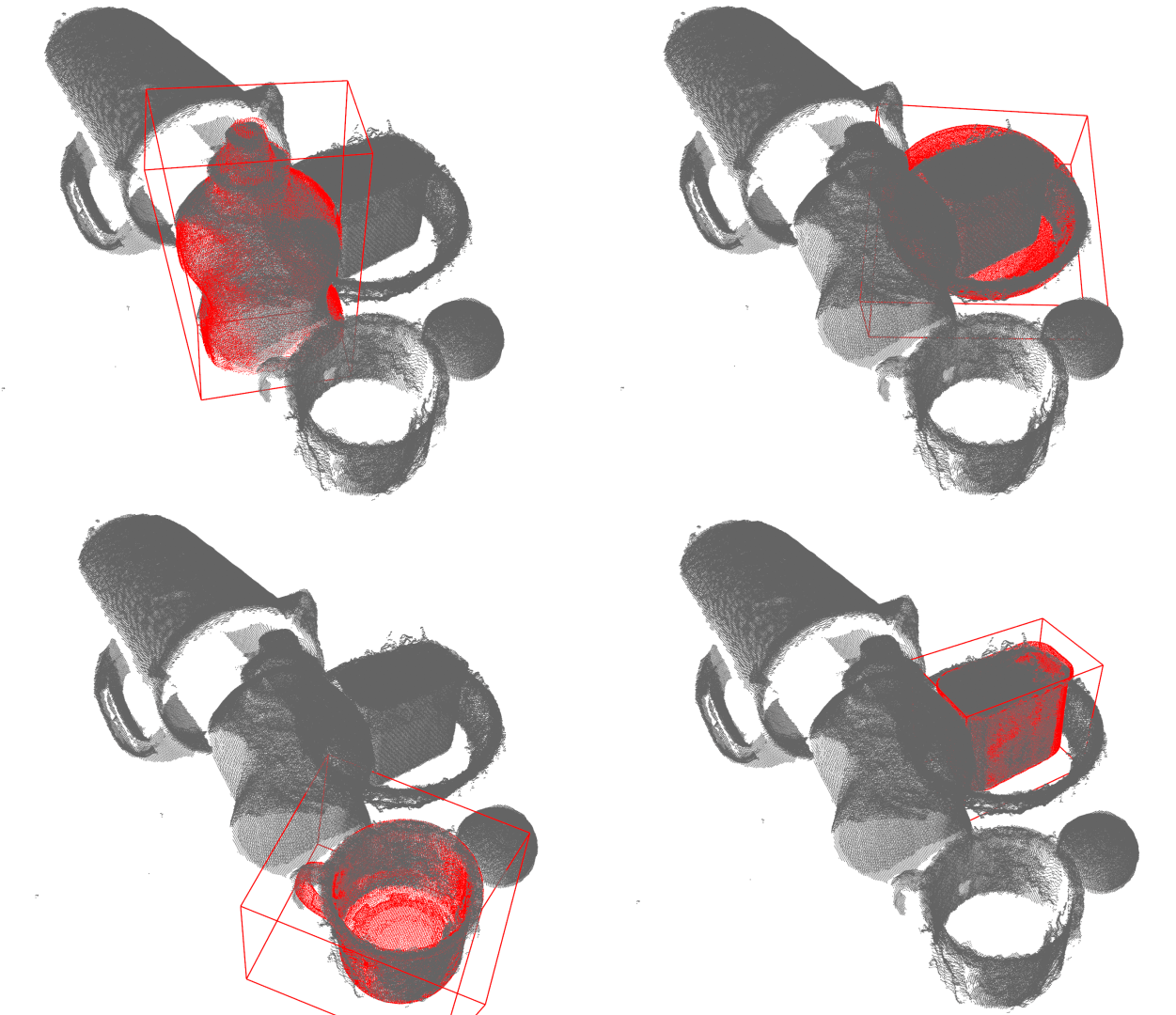}
    \caption{Sample images illustrating the 3D poses estimated for 4 different objects (mustard bottle, bowl, mug, potted meat can) using the proposed approach. Transformed models using estimated poses are shown in red.}
    \label{fig:pose_results}
\end{figure}
%
To validate the pose estimation method presented in Sec. \ref{sec:pose_estimation}, a scene with six randomly positioned objects on a table is constructed. Once the robot finishes generating the scene cloud, the operator selects an object from the terminal interface. Our method then identifies this object and estimates its pose. Fig. \ref{fig:pose_results} shows the results obtained for four objects, where the reference model (in red) is aligned with the scene cloud (in grey) using the estimated pose. The oriented bounding box of the transformed reference model, shown in red, is used to segment points from the scene that belong to the object. This segmented object region cloud is used for the grasp generation stage. As a side note, both the reference model and the scene clouds are expressed in the same reference frame. The reference frame of the model is notably located at its center of mass. It can be seen from the results that even though the scene is composed of multiple partially observed objects, the algorithm is able to locate and estimate the pose of the selected object.
%

%
%
\subsubsection{Grasp generation analysis}
%
%
\midsepremove
\begin{table*}
    \caption{Performance of the SpectGRASP in computing grasp hypotheses for various objects.}
    \label{tab:grasp_generation}
    \centering
    \begin{tabularx}{0.9\textwidth}{>{\raggedright\arraybackslash}X|
                              >{\centering\arraybackslash}X|
                              >{\centering\arraybackslash}X|
                             >{\centering\arraybackslash}X|
                             >{\centering\arraybackslash}X|
                             >{\centering\arraybackslash}X|
                             >{\centering\arraybackslash}X}
    \toprule
    \rowcolor[gray]{.9}
    Object name $\rightarrow$ &  \textbf{Bleach} & \textbf{Bowl} & \textbf{Mug} & \textbf{Screw Driver} & \textbf{Pitcher} & \textbf{Meat Can}\\
     \midrule
        Num grasps & 165833 &  471 & 5507 & 123742 & 6520 & 111719 \\
        \midrule
        Comp. time [s] & 135.049 & 3.08054 & 28.6959 & 61.9418 & 62.9619 & 91.8644 \\
     \bottomrule
\end{tabularx}
\end{table*}
\begin{figure}
    \centering
    \includegraphics[width=\columnwidth]{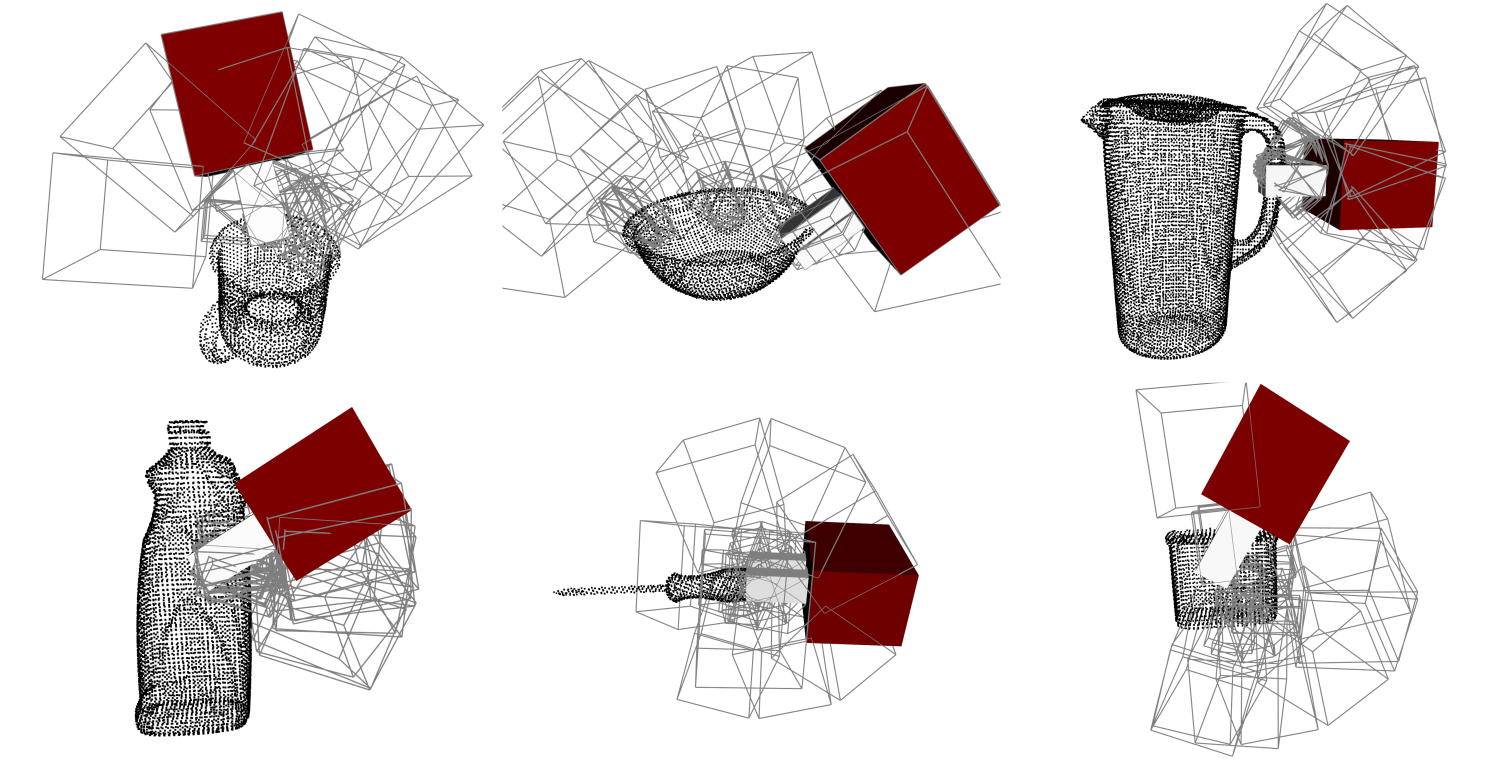}
    \caption{Grasps generated for test objects using SpectGRASP method. Red solid grasp is the top-ranked one and the the wireframes represent remaining grasp hypotheses. Single full object models are shown for easy understanding.}
    \label{fig:grasps_results}
\end{figure}
These tests are performed to demonstrate the grasp generation capability of the proposed pipeline, i.e., using our SpectGRASP \cite{adjigble2021spectgrasp}. Although the method is capable of generating grasps for multi-object scenes, single object clouds are used for these tests. This is because the grasps are generated specifically for the segmented region of a user selected object. Sample results for six different objects are shown in Fig. \ref{fig:grasps_results}. Out of the compiled grasps, only the top 10 hypotheses are shown in the figure with the rank-1 grasp in a solid red frame. The modular nature of SpectGRASP allows computing grasps without requiring the kinematic model of the robot, meaning that grasps are computed as if the hand is detached from the robot. At this generation stage of our method, the obtained grasps are not checked for kinematic feasibility or the physical reachability of the robot. Nonetheless, collision detection is still performed between the gripper and the object. Table \ref{tab:grasp_generation} shows the number of grasps computed and the time taken to compute them for four different objects. On average, 815.025 grasps are generated per second. These results clearly demonstrate the efficiency of SpectGRASP.

\subsubsection{Grasp re-ranking analysis}
\begin{figure}
    \centering
    \includegraphics[width=\columnwidth]{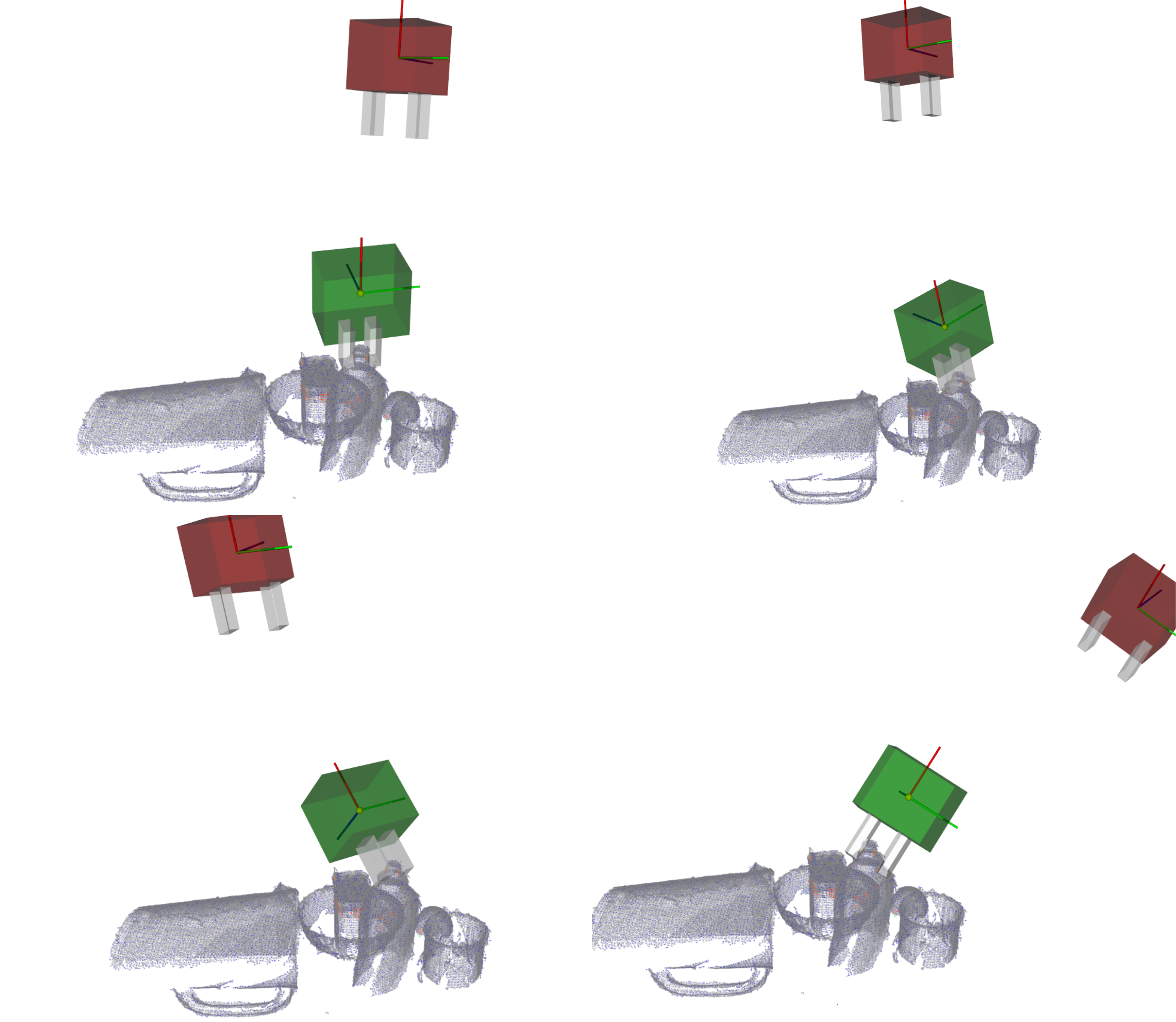}
    \caption{Illustration of the proposed re-ranking schema. Best grasps are dynamically updated based on the current pose of the robot hand.}
    \label{fig:rerank_results}
\end{figure}

%
The proposed re-ranking strategy is evaluated in this section. A scene composed of multiple randomly positioned objects is used. The re-ranking module is automatically activated once the grasps are generated for the user-selected object. The operator is able to teleoperate the robot using the haptic device. During manual teleoperation (without assistance), the closest 100 grasps to the current Cartesian position of the gripper are selected and re-ranked using \eqref{eq:re_ranking}. The grasp with the highest score is selected and displayed on the screen. Sample results for the ``mustard bottle" object are shown in Fig. \ref{fig:rerank_results}. The re-ranked best grasp (green gripper), is automatically updated based on the current position of the robot hand (red gripper).

\subsubsection{Force guidance analysis}
\begin{figure}
    \centering
    \includegraphics[width=\columnwidth]{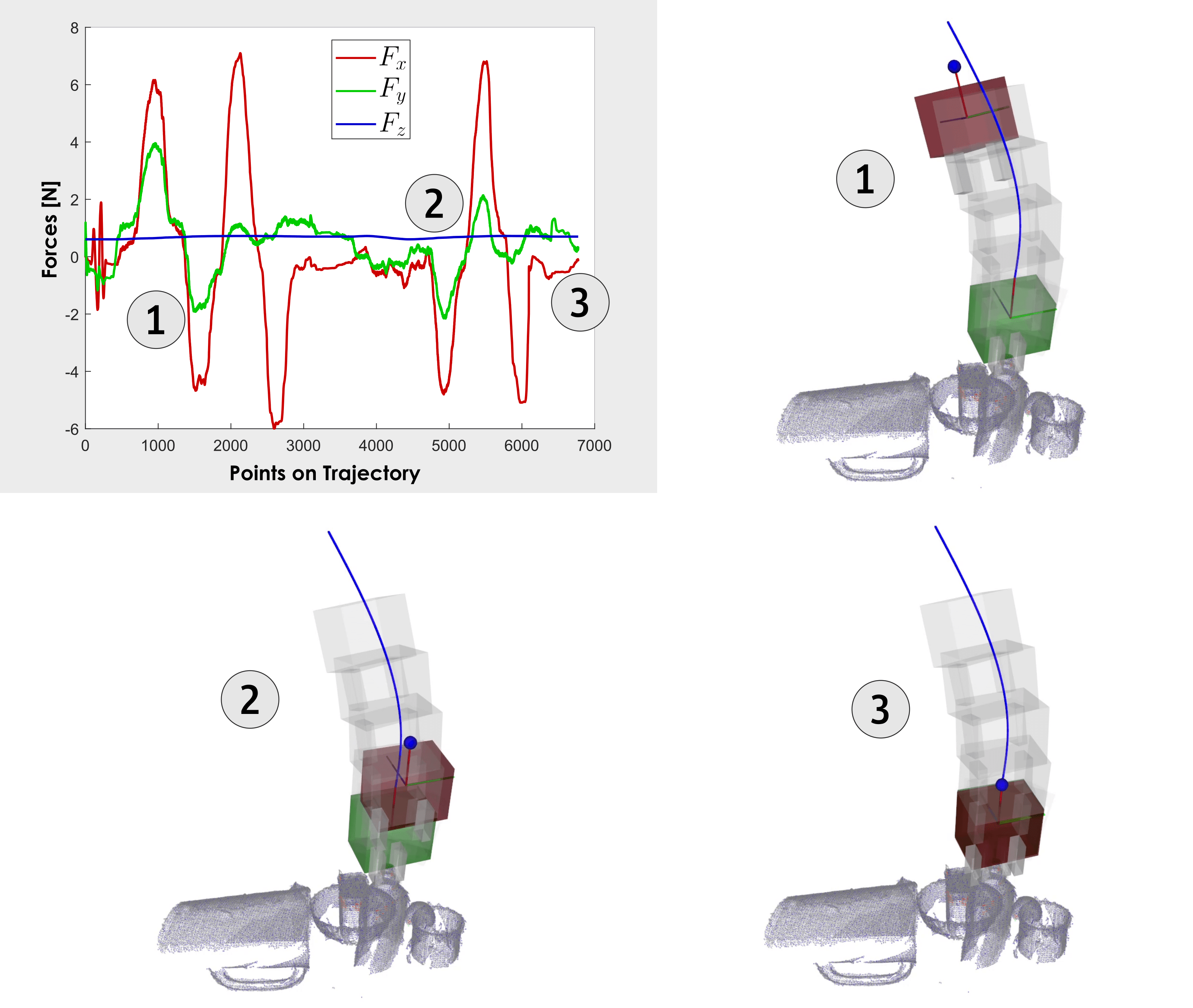}
    \caption{Illustration of the force guidance with automatic orientation alignment. Operator forces are high when moved away from blue path.}
    \label{fig:force_guide_results}
\end{figure}

The results of the force guidance experiment using the same scene as the previous experiment are illustrated in Fig. \ref{fig:force_guide_results}. The operator activates the force guidance by pressing a button on the haptic device. In the figure, the computed zero-force path trajectory is displayed in blue, while the interpolated poses of the gripper along the trajectory are shown as gray transparent grippers. These poses also illustrate the automatically computed orientations for the gripper. Any attempt to deviate from the path generates a force that tries to bring the operator back on the path. The zero-force trajectory notably simplifies the complex problem of reaching and grasping an object. 

\begin{figure*}
    \centering
    \includegraphics[width=0.85\textwidth]{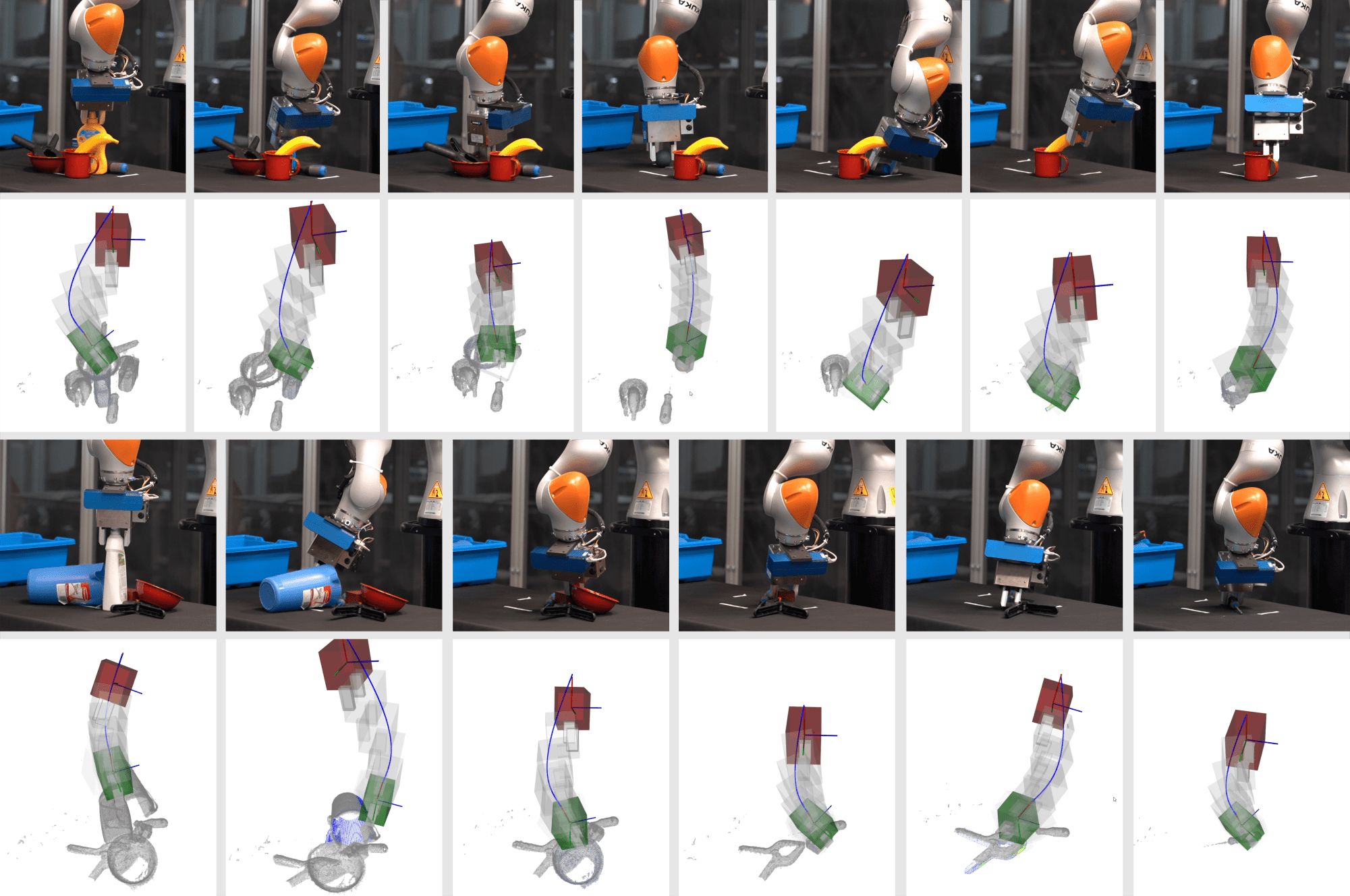}
    \caption{Illustration of clutter clearance with our proposed method. Two sample clutters and their assisted paths are shown.}
    \label{fig:clutter_results}
\end{figure*}
\subsection{Clutter clearance tests}\label{subsec:clutter_tests}
%
Three different clutter scenes are used to evaluate the heap clearance functionality of our method. Fig. \ref{fig:clutter_results} shows two of them being cleared. The operator is tasked with sequentially clearing the scene by selecting objects of their choice. The following steps are followed by the operator in accomplishing the task: %
\begin{enumerate*}[label=(\roman*)]
\item select the object of choice from the given list and input its ID in the interface;
\item select a target grasp (green gripper) of choice by moving the haptic device;
\item activate the guidance from the haptic interface and follow the zero-force-torque trajectory;
\item when reached the target location, close the gripper and manually teleoperate to the dropping location – blue bin in Fig. \ref{fig:setup_objects}a.
\end{enumerate*}
Based on the results, all objects in the three cluttered scenes were successfully handled on the first attempt. Since the object heaps are randomly generated with random number of objects (minimum 6), if the operator selects an object that is not present in the scene, the system returns a low alignment score and a warning. The operator can then choose another object, and the previous selection becomes unavailable. Also note that after every successful object removal, the scene point cloud is regenerated. Detailed results can be seen in the video at \url{https://youtu.be/SqDwjwpluc4}.

%
\section{Conclusion}
\label{sec:conclusion}
This paper presents a telemanipulation approach for grasping desired objects from cluttered scenes. Our method combines a 3D pose estimator with a dynamic grasp re-ranking strategy to identify the best grasp candidate in real-time, based on the gripper's current pose. Using our SpectGRASP method, we generate grasps efficiently for a wide variety of objects. The zero-force trajectory enables the operator to seamlessly grasp the desired object by following a reference trajectory, while the robot's orientation is automatically controlled. Experiments performed on different scenes demonstrate that our pipeline can efficiently remove all objects from a heap. In future work, we aim to provide a more natural way to control the robot using virtual reality and integrate a multi-finger robotic hand for dexterous manipulation.
\bibliographystyle{IEEEtran}
\bibliography{references}

\end{document}